\documentclass[11pt,a4paper]{article}
\usepackage[hyperref]{naaclhlt2018}
\usepackage{times}
\usepackage{latexsym}
\usepackage{amsmath}
\usepackage{url}
\usepackage{subcaption}
\usepackage{graphicx}

\usepackage{caption}
\usepackage{multirow}
\usepackage{amssymb}
\usepackage{mathrsfs}
\usepackage{graphics}
\usepackage{graphicx}
\usepackage{ifsym}
\usepackage{amsmath}
\usepackage{subcaption}
\usepackage{booktabs}
\usepackage{multirow}
\usepackage{array}
\usepackage{bbm}
\usepackage{multicol}
\usepackage{blindtext}
\usepackage{algorithm}
\usepackage{algpseudocode}
\usepackage{tikz}

\usepackage{lipsum}
\usepackage{ntheorem}
\usepackage{makecell}

\DeclareMathOperator*{\argmax}{arg\,max}

\DeclareCaptionFont{10pt}{\fontsize{10pt}{12pt}\selectfont}
\aclfinalcopy 

\title{A Bi-model based RNN Semantic Frame Parsing Model for Intent Detection and Slot Filling}

\author{Yu Wang \\
  \\\And
  Yilin Shen \\
  Samsung Research America  \\
  {\tt $\lbrace$yu.wang1, yilin.shen, hongxia.jin$\rbrace$@samsung.com} \\
  \\\And
  Hongxia Jin \\
  }

\date{}

\begin{document}
\maketitle
\begin{abstract}
Intent detection and slot filling are two main tasks for building a spoken language understanding(SLU) system. Multiple deep learning based models have demonstrated good results on these tasks . The most effective algorithms are based on the structures of sequence to sequence models (or "encoder-decoder" models), and generate the intents and semantic tags either using separate models(\cite{yao2014spoken,mesnil2015using,peng2015recurrent,kurata2016leveraging,hahn2011comparing}) or a joint model (\cite{liu2016attention,hakkani2016multi, guo2014joint}). Most of the previous studies, however, either treat the intent detection and slot filling as two separate parallel tasks, or use a sequence to sequence model to generate both semantic tags and intent. Most of these approaches use one (joint) NN based model (including encoder-decoder structure) to model two tasks, hence may not fully take advantage of the cross-impact between them. In this paper, new Bi-model based RNN semantic frame parsing network structures are designed to perform the intent detection and slot filling tasks jointly, by considering their cross-impact to each other using two correlated bidirectional LSTMs (BLSTM). Our Bi-model structure with a decoder achieves state-of-the-art result on the benchmark ATIS data \cite{hemphill1990atis,tur2010left}, with about 0.5$\%$ intent accuracy improvement and 0.9 $\%$ slot filling improvement.
\end{abstract}

\section{Introduction}
The research on spoken language understanding (SLU) system has progressed extremely fast during the past decades. Two important tasks in an SLU system are intent detection and slot filling. These two tasks are normally considered as parallel tasks but may have cross-impact on each other. The intent detection is treated as an utterance classification problem, which can be modeled using conventional classifiers including regression, support vector machines (SVMs) or even deep neural networks \cite{haffner2003optimizing, sarikaya2011deep}.
The slot filling task can be formulated as a sequence labeling problem, and the most popular approaches with good performances are using conditional random fields (CRFs) and recurrent neural networks (RNN) as recent works \cite{xu2013convolutional}. 

Some works also suggested using one joint RNN model for generating results of the two tasks together, by taking advantage of the sequence to sequence\cite{sutskever2014sequence} (or encoder-decoder) model, which also gives decent results as in literature\cite{liu2016attention}.

In this paper, Bi-model based RNN structures are proposed to take the cross-impact between two tasks into account, hence can further improve the performance of modeling an SLU system. These models can generate the intent and semantic tags concurrently for each utterance. In our Bi-model structures, two task-networks are built for the purpose of intent detection and slot filling. Each task-network includes one BLSTM with or without a LSTM decoder \cite{hochreiter1997long,graves2005framewise}.

The paper is organized as following: In section 2, a brief overview of existing deep learning approaches for intent detection and slot fillings are given. The new proposed Bi-model based RNN approach will be illustrated in detail in section 3. In section 4, two experiments on different datasets will be given. One is performed on the ATIS benchmark dataset, in order to demonstrate a state-of-the-art result for both semantic parsing tasks. The other experiment is tested on our internal multi-domain dataset by comparing our new algorithm with the current best performed RNN based joint model in literature for intent detection and slot filling.
\section{Background}
In this section, a brief background overview on using deep learning and RNN based approaches to perform intent detection and slot filling tasks is given. The joint model algorithm is also discussed for further comparison purpose.
\subsection{Deep neural network for intent detection}
Using deep neural networks for intent detection is similar to a standard classification problem, the only difference is that this classifier is trained under a specific domain. For example, all data in ATIS dataset is under the flight reservation domain with 18 different intent labels. There are mainly two types of models that can be used: one is a feed-forward model by taking the average of all words' vectors in an utterance as its input, the other way is by using the recurrent neural network which can take each word in an utterance as a vector one by one \cite{xu2014contextual}. 
\subsection{Recurrent Neural network for slot filling}
The slot filling task is a bit different from intent detection as there are multiple outputs for the task, hence only RNN model is a feasible approach for this scenario. The most straight-forward way is using single RNN model generating multiple semanctic tags sequentially by reading in each word one by one \cite{liu2015recurrent,mesnil2015using,peng2015recurrent}. This approach has a constrain that the number of slot tags generated should be the same as that of words in an utterance. Another way to overcome this limitation is by using an encoder-decoder model containing two RNN models as an encoder for input and a decoder for output \cite{liu2016attention}. The advantage of doing this is that it gives the system capability of matching an input utterance and output slot tags with different lengths without the need of alignment. Besides using RNN, It is also possible to use the convolutional neural network (CNN) together with a conditional random field (CRF) to achieve slot filling task \cite{xu2013convolutional}.
\subsection{Joint model for two tasks}
It is also possible to use one joint model for intent detection and slot filling \cite{guo2014joint, liu2016attention,liu2016joint,zhang2016joint,hakkani2016multi}. One way is by using one encoder with two decoders, the first decoder will generate sequential semantic tags and the second decoder generates the intent. Another approach is by consolidating the hidden states information from an RNN slot filling model, then generates its intent using an attention model \cite{liu2016attention}. Both of the two approaches demonstrates very good results on ATIS dataset. 
\section{Bi-model RNN structures for joint semantic frame parsing }
Despite the success of RNN based sequence to sequence (or encoder-decoder) model on both tasks, most of the approaches in literature still use one single RNN model for each task or both tasks. They treat the intent detection and slot filling as two separate tasks. In this section, two new Bi-model structures are proposed to take their cross-impact into account, hence further improve their performance. One structure takes the advantage of a decoder structure and the other doesn't. An asynchronous training approach based on two models' cost functions is designed to adapt to these new structures. 
\subsection{Bi-model RNN Structures}
\begin{figure}
\begin{subfigure}{.5\textwidth}
  \centering
  \includegraphics[width=0.9\linewidth]{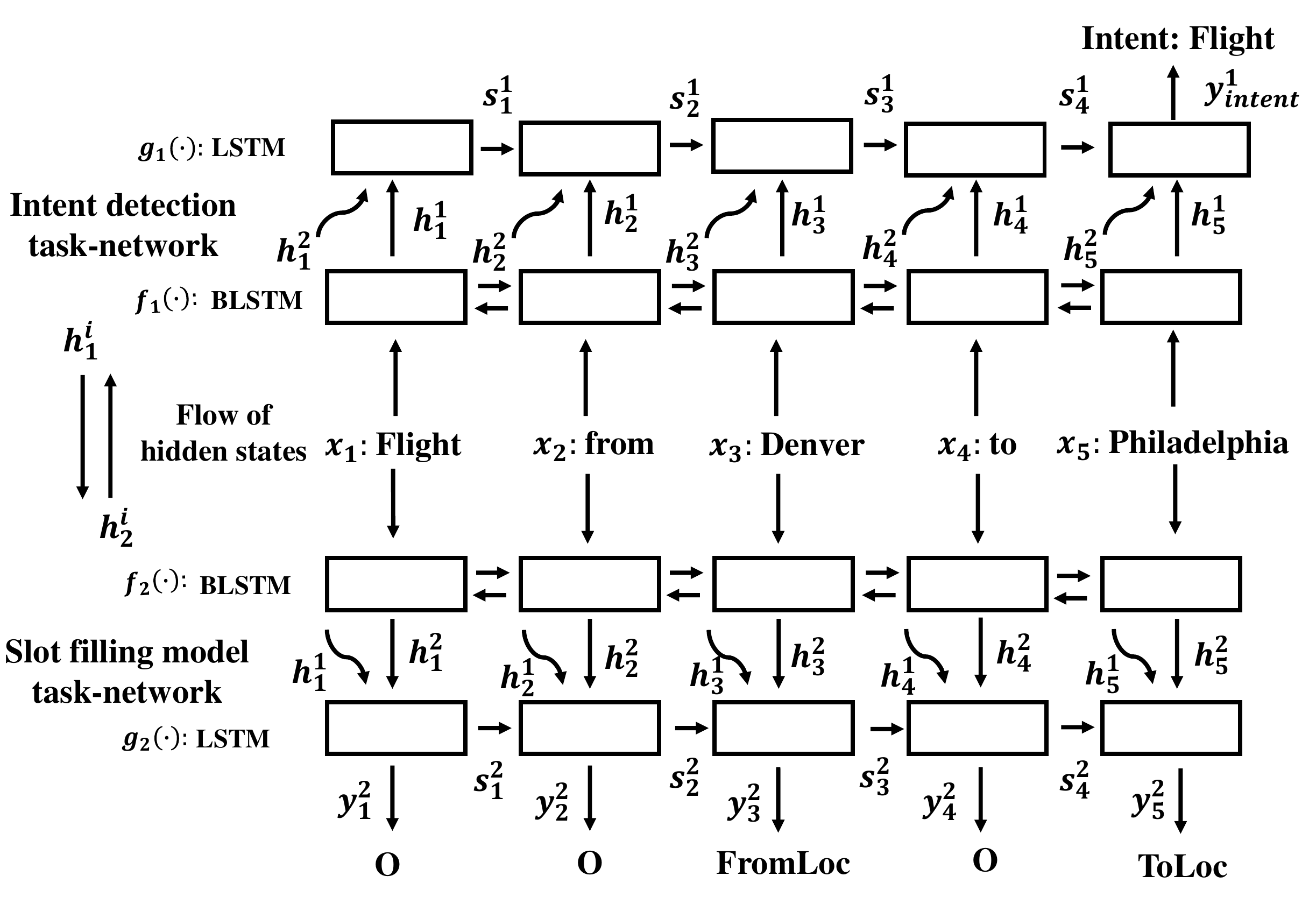}
  \caption{Bi-model structure with a decoder}
  \label{bimodel1}
\end{subfigure}%
\hfill
\begin{subfigure}{.5\textwidth}
  \centering
  \includegraphics[width=0.9\linewidth]{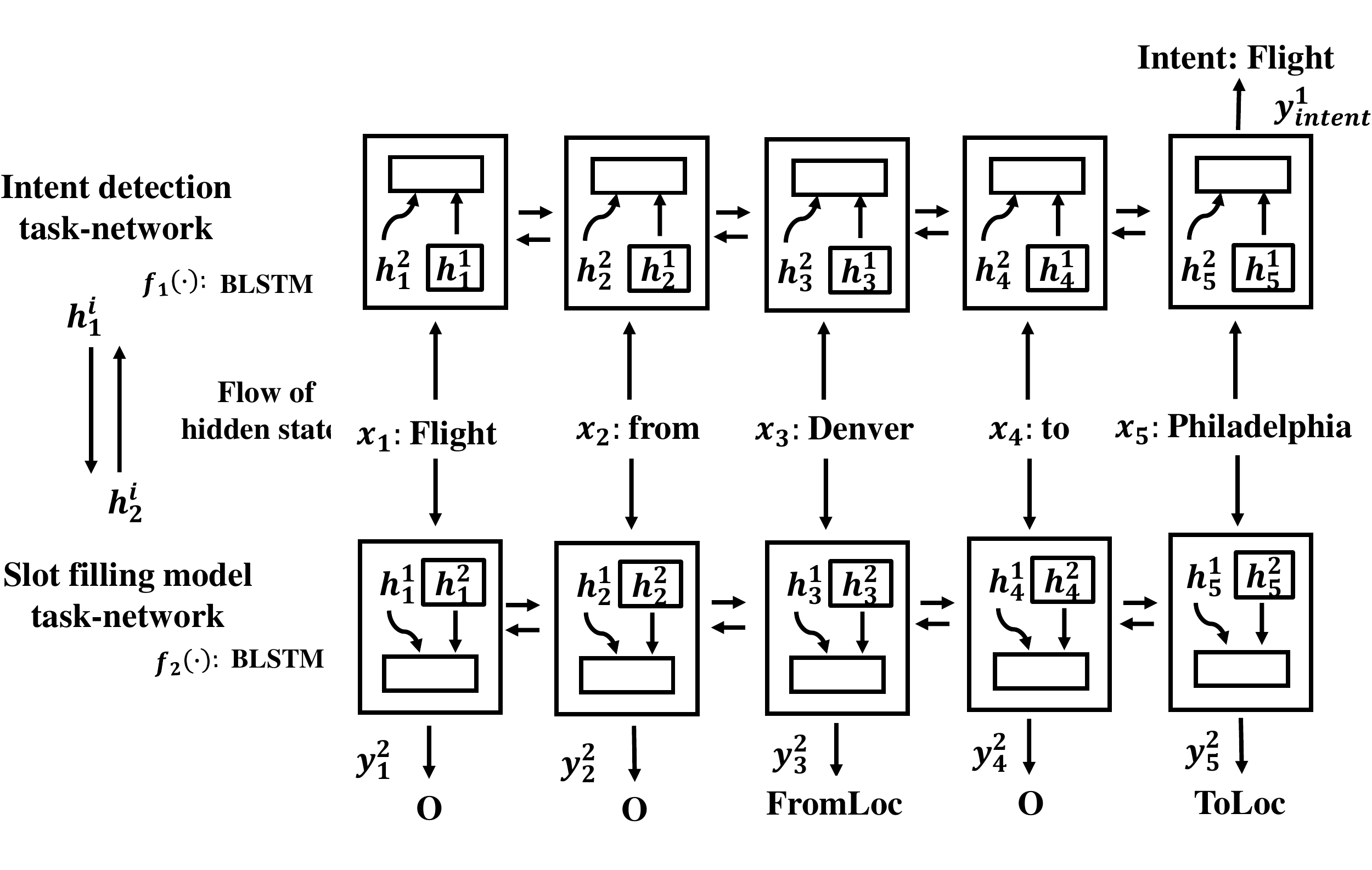}
  \caption{Bi-model structure without a decoder}
  \label{bimodel2}
\end{subfigure}
\caption{Bi-model structure}
\label{bimodel}
\end{figure}

A graphical illustration of two Bi-model structures with and without a decoder is shown in Figure \ref{bimodel}. The two structures are quite similar to each other except that Figure \ref{bimodel1} contains a LSTM based decoder, hence there is an extra decoder state $s_t$ to be cascaded besides the encoder state $h_t$. 
\\
{\it{Remarks:}\\
The concept of using information from multiple-model/multi-modal to achieve better performance has been widely used in deep learning  \cite{dean2012large, wang2017new,ngiam2011multimodal,srivastava2012multimodal}, system identification  \cite{murray1997multiple,narendra2014stability, narendra2015extension} and also  reinforcement learning field recently \cite{narendra2016fast,wang2018Boost}. Instead of using collective information, in this paper, our work introduces a totally new approach of training multiple neural networks asynchronously by sharing their internal state information.}

\subsubsection{Bi-model structure with a decoder}
The Bi-model structure with a decoder is shown as in Figure \ref{bimodel1}. There are two inter-connected bidirectional LSTMs (BLSTMs) in the structure, one is for intent detection and the other is for slot filling. Each BLSTM reads in the input utterance sequences $(x_1,x_2,\cdots,x_n)$ forward and backward, and generates two sequences of hidden states $hf_t$ and $hb_t$. A concatenation of $hf_t$ and $hb_t$ forms a final BLSTM state $h_t=[hf_t,hb_t]$ at time step $t$. Hence, Our bidirectional LSTM $f_i(\cdot)$ generates a sequence of hidden states $(h_1^i, h_2^i,\cdots, h_n^i)$, where $i=1$ corresponds the network for intent detection task and $i=2$ is for the slot filling task. 

In order to detect intent, hidden state $h_t^1$ is combined together with $h_t^2$ from the other bidirectional LSTM $f_2(\cdot)$ in slot filling task-network to generate the state of $g_1(\cdot)$, $s_t^1$, at time step $t$:
\begin{equation}
\begin{split}
s_t^1&=\phi(s_{t-1}^1,h^1_{n-1},h^2_{n-1})\\
y_{intent}^1&=\argmax_{\hat{y}_n^{1}} P(\hat{y}_n^{1}|s_{n-1}^1,h^1_{n-1},h^2_{n-1})
\end{split}
\label{RNNState}
\end{equation}
where $\hat{y}_n^{1}$ contains the predicted probabilities for all intent labels at the last time step $n$.

For the slot filling task, a similar network structure is constructed with a BLSTM $f_2(\cdot)$ and a LSTM $g_2(\cdot)$. $f_2(\cdot)$ is the same as $f_1(\cdot)$, by reading in the a word sequence as its input. The difference is that there will be an output $y_t^2$ at each time step $t$ for $g_2(\cdot)$, as it is a sequence labeling problem. At each step $t$:
\begin{equation}
\begin{split}
s^2_t&=\psi(h^2_{t-1},h^1_{t-1},s^2_{t-1},y_{t-1}^2)\\
y_t^2&=\argmax_{\hat{y}_t^{2}} P(\hat{y}_t^{2}|h^1_{t-1},h^2_{t-1},s^2_{t-1},y_{t-1}^2)
\end{split}
\label{RNN2State}
\end{equation}
where $y_t^2$ is the predicted semantic tags at time step $t$.
\subsubsection{Bi-Model structure without a decoder}
The Bi-model structure without a decoder is shown as in Figure \ref{bimodel2}. In this model, there is no LSTM decoder as in the previous model. 

For the intent task, only one predicted output label $y_{intent}^1$ is generated from BLSTM $f_1(\cdot)$ at the last time step $n$, where $n$ is the length of the utterance. Similarly, the state value $h_t^1$ and output intent label are generated as:
\begin{equation}
\begin{split}
h^1_t&=\phi(h^1_{t-1},h^2_{t-1})\\
y_{intent}^1&=\argmax_{\hat{y}_n^{1}} P(\hat{y}_n^{1}|h^1_{n-1},h^2_{n-1})
\end{split}
\end{equation}
For the slot filling task, the basic structure of BLSTM $f_2(\cdot)$ is similar to that for the intent detection task $f_1(\cdot)$, except that there is one slot tag label $y_t^2$ generated at each time step $t$. It also takes the hidden state from two BLSTMs $f_1(\cdot)$ and $f_2(\cdot)$, \emph{i.e.} $h_{t-1}^1$ and $h_{t-1}^2$, plus the output tag $y_{t-1}^2$ together to generate its next state value $h_t^2$ and also the slot tag $y_{t}^2$. To represent this as a function mathematically:
\begin{equation}
\begin{split}
h^2_t&=\psi(h^2_{t-1},h^1_{t-1},y_{t-1}^2)\\
y_t^2&=\argmax_{\hat{y}_t^{2}} P(\hat{y}_t^{2}|h^1_{t-1},h^2_{t-1},y_{t-1}^2)
\end{split}
\end{equation}
\subsubsection{Asynchronous training}
One of the major differences in the Bi-model structure is its asynchronous training, which trains two task-networks based on their own cost functions in an asynchronous manner. The loss function for intent detection task-network is $\mathcal{L}_1$, and for slot filling is $\mathcal{L}_2$. $\mathcal{L}_1$ and $\mathcal{L}_2$ are defined using cross entropy as:
\begin{equation}
\mathcal{L}_1\triangleq-\sum_{i=1}^k\hat{y}_{intent}^{1,i}\log(y_{intent}^{1,i})
\end{equation}
and 
\begin{equation}
\mathcal{L}_2\triangleq-\sum_{j=1}^n\sum_{i=1}^m\hat{y}_{j}^{2,i}\log(y_{j}^{2,i})
\end{equation}
where $k$ is the number of intent label types, $m$ is the number of semantic tag types and $n$ is the number of words in a word sequence.
In each training iteration, both intent detection and slot filling networks will generate a groups of hidden states $h^1$ and $h^2$ from the models in previous iteration. The intent detection task-network reads in a batch of input data $x_i$ and hidden states $h^2$, and generates the estimated intent labels $\hat{y}_{intent}^{1}$. The intent detection task-network computes its cost based on function $\mathcal{L}_1$ and trained on that. Then the same batch of data $x_i$ will be fed into the slot filling task-network together with the hidden state $h^1$ from intent task-network, and further generates a batch of outputs $y_i^2$ for each time step. Its cost value is then computed based on cost function $\mathcal{L}_2$, and further trained on that.

The reason of using asynchronous training approach is because of the importance of keeping two separate cost functions for different tasks. Doing this has two main advantages:\\
1. It filters the negative impact between two tasks in comparison to using only one joint model, by capturing more useful information and overcoming the structural limitation of one model.\\
2. The cross-impact between two tasks can only be learned by sharing hidden states of two models, which are trained using two cost functions separately. 
\section{Experiments}
In this section, our new proposed Bi-model structures are trained and tested on two datasets, one is the public ATIS dataset \cite{hemphill1990atis} containing audio recordings of flight reservations, and the other is our self-collected datset in three different domains: Food, Home and Movie. The ATIS dataset used in this paper follows the same format as in \cite{liu2015recurrent, mesnil2015using,xu2013convolutional, liu2016attention}. The training set contains 4978 utterance and the test set contains 893 utterance, with a total of 18 intent classes and 127 slot labels. The number of data for our self-collected dataset will be given in the corresponding experiment sections with a more detailed explanation.
The performance is evaluated based on the classification accuracy for intent detection task and F1-score for slot filling task.
\subsection{Training Setup}
The layer sizes for both the LSTM and BLSTM networks in our model are chosen as 200. Based on the size of our dataset, the number of hidden layers is chosen as 2 and Adam optimization is used as in \cite{kingma2014adam}. The size of word embedding is 300, which are initialized randomly at the beginning of experiment.
\subsection{Performance on the ATIS dataset}
Our first experiment is conducted on the ATIS benchmark dataset, and compared with the current existing approaches, by evaluating their intent detection accuracy and slot filling F1 scores.
\begin{table}[ht]\scriptsize
\centering

	\begin{tabular}{>{\centering\arraybackslash}p{3.5cm}|>{\centering\arraybackslash}p{1.5cm}>{\centering\arraybackslash}p{1.5cm}}
		\toprule
		
		\multirow{1}{*}{\textbf{Model}} & \multirow{1}{*}{\makecell{\textbf{F1 Score}}} & \multirow{1}{*}{\makecell{\textbf{Intent Accuracy}}}\\
		\midrule
		\midrule
		\multirow{2}{*} {}Recursive NN  & 93.96\% & 95.4\%  \\
		\cite{guo2014joint}\\ 
		\multirow{2}{*}{}Joint model with recurrent intent and slot label context &94.47\% & 98.43\%\\
		\cite{liu2016joint}\\
		\multirow{2}{*}{}Joint model with recurrent slot label context &94.64\% & 98.21\%\\
		\cite{liu2016joint}\\
		\multirow{2}{*}{}RNN with Label Sampling  & 94.89\% & NA \\
		\cite{liu2015recurrent}\\
		\multirow{2}{*}{}Hybrid RNN & 95.06\% & NA  \\
		\cite{mesnil2015using}\\		
		\multirow{2}{*}{}RNN-EM  & 95.25\% & NA  \\
		\cite{peng2015recurrent}\\
		\multirow{2}{*}{}CNN CRF & 95.35\% & NA  \\
		\cite{xu2013convolutional}\\		
		\multirow{2}{*}{}Encoder-labeler Deep LSTM & 95.66\% & NA  \\
		\cite{kurata2016leveraging}\\
		\multirow{2}{*}{}Joint GRU Model (W) &95.49\%&98.10\%\\
		\cite{zhang2016joint}\\
		\multirow{2}{*}{}Attention Encoder-Decoder NN &95.87\% & 98.43\%\\
		\cite{liu2016attention}\\
		\multirow{2}{*}{}Attention BiRNN & 95.98\% & 98.21\%\\
		\cite{liu2016attention}\\
		\midrule
		\multirow{2}{*}{}Bi-model without a decoder & \textbf{96.65\%} & \textbf{98.76}\%\\
		
		\multirow{2}{*}{}Bi-model with a decoder & \textbf{96.89\%} & \textbf{98.99}\%\\

		\bottomrule		
	\end{tabular}
	\caption{Performance of Different Models on ATIS Dataset}
	\label{table:data_comparison}
\end{table}
A detailed comparison is given in Table \ref{table:data_comparison}. Some of the models are designed for single slot filling task, hence only F1 scores are given. It can be observed that the new proposed Bi-model structures outperform the current state-of-the-art results on both intent detection and slot filling tasks, and the Bi-model with a decoder also outperform that without a decoder on our ATIS dataset. The current Bi-model with a decoder shows the state-of-the-art performance on ATIS benchmark dataset with 0.9$\%$ improvement on F1 score and 0.5$\%$ improvement on intent accuracy.\\
{\it{Remarks:}\\
1.  It is worth noticing that the complexities of encoder-decoder based models are normally higher than the models without using encoder-decoder structures, since two networks are used and more parameters need to be updated. This is another reason why we use two models with/without using encoder-decoder structures to demonstrate the new bi-model structure design. It can also be observed that the model with a decoder gives a better result due to its higher complexity.\\
2. It is also shown in the table that the joint model in \cite{ liu2015recurrent, liu2016attention} achieves better performance on intent detection task with slight degradation on slot filling, so a joint model is not necessary always better for both tasks. The bi-model approach overcomes this issue by generating two tasks' results separately.\\
3. Despite the absolute improvement of intent accuracy and F1 scores are only  0.5$\%$ and 0.9$\%$ on ATIS dataset, the relative improvement is not small. For intent accuracy, the number of wrongly classified utterances in test dataset reduced from 14 to 9, which gives us the 35.7$\%$ relative improvement on intent accuracy. Similarly, the relative improvement on F1 score is 22.63$\%$.\\}

\subsection{Performance on multi-domain data}
In this experiment, the Bi-model structures are further tested on an internal collected dataset from our users in three domains: food, home and movie. There are 3 intents for each domain, 15 semantic tags in food domain, 16 semantic tags in home domain, 14 semantic tags in movie domain.  The data size of each domain is listed as in Table \ref{table:data_comparison2}, and the split is 70$\%$ for training, 10$\%$ for validation and 20$\%$ for test.

Due to the space limitation, only the best performed semantic frame parsing model on ATIS dataset in literature,i.e. attention based BiRNN \cite{liu2016attention} is used for comparison with our Bi-model structures.
\begin{table}[ht]\scriptsize

\centering

	\begin{tabular}{>{\centering\arraybackslash}p{0.8cm}|>{\centering\arraybackslash}p{2.6cm}|c>{\centering\arraybackslash}p{0.4cm}>{\centering\arraybackslash}p{1.5cm}c}
		\toprule
		\multirow{1}{*}{\textbf{Domain}} & \textbf{\multirow{1}{*}{\makecell{\textbf{SLU model}}}} & \multirow{1}{*}{\makecell{\textbf{Size}}} & \multirow{1}{*}{}{\textbf{F1 Score}}&\multirow{1}{*}{\makecell{\textbf{Accuracy}}}\\
		\midrule
		\multirow{3}{*}{Movie} & Attention BiRNN & 979 & 92.1\%& 92.86\%  \\
		& Bi-model without a decoder & 979  & 93.3\%& 94.89\% \\
		& Bi-model with a decoder & 979 & \textbf{93.8\%}& \textbf{95.91\%}  \\
		\midrule
		\multirow{3}{*}{Food} & Attention BiRNN & 983 & 92.3\%& 98.48\%  \\
		& Bi-model without a decoder & 983  & 93.6\%& 98.98\% \\
		& Bi-model with a decoder & 983  & \textbf{95.8\%}& \textbf{99.49\%} \\
		\midrule
		\multirow{3}{*}{Home} & Attention BiRNN & 689 & 96.5\%& 97.83\%  \\
		& Bi-model without a decoder & 689  & 97.8\%& 98.55\% \\
		& Bi-model with a decoder & 689& \textbf{98.2\%}& \textbf{99.27\%}\\
		\bottomrule		
	\end{tabular}
	\caption{Performance Comparison between Bi-model Structures and Attention BiRNN}
	\label{table:data_comparison2}
\end{table}
Table \ref{table:data_comparison2} shows a performance comparison in three domains of data. The Bi-model structure with a decoder gives the best performance in all cases based on its intent accuracy and slot filling F1 score. The intent accuracy has at least 0.5·$\%$ improvement,  the F1 score improvement is around 1$\%$ to 3$\%$ for different domains.

\section{Conclusion}
In this paper, a novel Bi-model based RNN semantic frame parsing model for intent detection and slot filling is proposed and tested. Two substructures are discussed with the help of a decoder or not. The Bi-model structures achieve state-of-the-art performance for both intent detection and slot filling on ATIS benchmark data, and also surpass the previous best SLU model on the multi-domain data. The Bi-model based RNN structure with a decoder also outperforms the Bi-model structure without a decoder on both ATIS and multi-domain data.

\bibliography{naaclhlt2018}
\bibliographystyle{acl_natbib}


\end{document}